\pdfoutput=1
\documentclass[11pt,a4paper]{article}
\usepackage[hyperref]{AACL-IJCNLP2020}
\usepackage{times}
\usepackage{latexsym}

\usepackage{bbm}
\usepackage{graphicx}
\usepackage{amsmath}
\usepackage{amsfonts}
\usepackage{boldline}
\usepackage{natbib}
\usepackage{caption}
\usepackage{tikz-dependency}
\usepackage{wrapfig}
\usepackage{multirow}
\usepackage{pgfplots}
\usepackage{filecontents}
\usepackage[T1]{fontenc}
\usepackage{subfiles}
\usepackage{readarray}
\usepackage{xcolor}
\usepackage{import}

\usepackage{microtype}

\aclfinalcopy 

\newcommand{\wvec}{\mathbf{w}}

\title{Second-Order Neural Dependency Parsing with Message Passing and End-to-End Training}

\author{Xinyu Wang \and Kewei Tu\thanks{\hspace{1mm} Kewei Tu is the corresponding author. } \\
 School of Information Science and Technology, ShanghaiTech University \\
 Shanghai Engineering Research Center of Intelligent Vision and Imaging \\
 Shanghai Institute of Microsystem and Information Technology, Chinese Academy of Sciences \\
 University of Chinese Academy of Sciences \\
  {\tt \{wangxy1,tukw\}@shanghaitech.edu.cn} \\
 }

\date{}

\begin{document}
\maketitle
\begin{abstract}
In this paper, we propose second-order graph-based neural dependency parsing using message passing and end-to-end neural networks. We empirically show that our approaches match the accuracy of very recent state-of-the-art second-order graph-based neural dependency parsers and have significantly faster speed in both training and testing. We also empirically show the advantage of second-order parsing over first-order parsing and observe that the usefulness of the head-selection structured constraint vanishes when using BERT embedding.
\end{abstract}

\section{Introduction}
Graph-based dependency parsing is a popular approach to dependency parsing that scores parse components of a sentence and then finds the highest scoring tree through inference. First-order graph-based dependency parsing takes individual dependency edges as the components of a parse tree, while higher-order dependency parsing considers more complex components consisting of multiple edges. There exist both exact inference algorithms \cite{carreras2007experiments,koo-collins-2010-efficient,ma-zhao-2012-fourth} and approximate inference algorithms \cite{mcdonald2006online,smith-eisner-2008-dependency,gormley-etal-2015-approximation} to find the best parse tree.
Recent work focused on neural network based graph dependency parsers \cite{kiperwasser-goldberg-2016-simple,wang-chang-2016-graph,cheng-etal-2016-bi,kuncoro-etal-2016-distilling,ma-hovy-2017-neural,dozat2016deep}. \citet{dozat2016deep} proposed a first-order graph-based neural dependency parsing approach with a simple head-selection training objective. It uses a biaffine function to score dependency edges and has high efficiency and good performance. Subsequent work introduced second-order inference into their parser. \citet{ji-etal-2019-graph} proposed a graph neural network that captures second-order information in token representations, which are then used for first-order parsing. Very recently, \citet{zhang2020efficient} proposed an efficient second-order tree CRF model for dependency parsing and achieved state-of-the-art performance.


In this paper, we first show how a previously proposed second-order semantic dependency parser \cite{wang-etal-2019-second} can be applied to syntactic dependency parsing with simple modifications. The parser is an end-to-end neural network derived from message passing inference on a conditional random field that encodes the second-order parsing problem. We then propose an alternative conditional random field that incorporates the head-selection constraint of syntactic dependency parsing, and derive a novel second-order dependency parser. We empirically compare the two second-order approaches and the first-order baselines on English Penn Tree Bank 3.0 (PTB), Chinese Penn Tree Bank 5.1 (CTB) and datasets of 12 languages in Universal Dependencies (UD). We show that our approaches achieve state-of-the-art performance on both PTB and CTB and our approaches are significantly faster than recently proposed second-order parsers. 

We also make two interesting observations from our empirical study. First, it is a common belief that contextual word embeddings such as ELMo \cite{peters-etal-2018-deep} and BERT \cite{devlin-etal-2019-bert} already conveys sufficient high-order information that renders high-order parsing less useful, but we find that second-order decoding is still helpful even with strong contextual embeddings like BERT. Second, while \citet{zhang-etal-2019-empirical} previously found that incoperating the head-selection constraint is helpful in first-order parsing, we find that with a better loss function design and hyper-parameter tuning both first- and second-order parsers without the head-selection constraint can match the accuracy of parsers with the head-selection constraint and can even outperform the latter when using BERT embedding.

Our approaches are closely related to the work of \citet{gormley-etal-2015-approximation}, which proposed a non-neural second-order parser based on Loopy Belief Propagation (LBP). Our work differs from theirs in that: 1) we use Mean Field Variational Inference (MFVI) instead of LBP, which \citet{wang-etal-2019-second} found is faster and equally accurate in practice; 2) we add the head-selection constraint and do not include the global tree constraint that is shown to produce only slight improvement \cite{zhang-etal-2019-empirical} but would complicate our neural network design and implementation; 3) we employ modern neural encoders and achieve much better parsing accuracy. Our approaches are also closely related to the very recent work of \citet{turbo2020}. The main difference is that we use MFVI while they use the dual decomposition algorithm $\text{AD}^\text{3}$ \citep{martins-etal-2011-dual,martins-etal-2013-turning} for approximate inference.

\section{Approach}
\citet{zhang-etal-2019-empirical} categorized different kinds of graph-based dependency parsers based on their structured output constraints according to the normalization for output scores. A \textbf{Local} approach views dependency parsing as a head-selection problem, in which each word selects exactly one dependency head. A \textbf{Single} approach places no structured constraint, viewing the existence of each possible dependency edge as an independent binary classification problem. 

The second-order semantic dependency parser of \citet{wang-etal-2019-second} is an end-to-end neural network derived from message passing inference on a conditional random field that encodes the second-order parsing problem. It is clearly a \textbf{Single} approach because of the lack of structured constraints in semantic dependency parsing. We can apply this approach to syntactic dependency parsing with two minor modifications. First, co-parents, one of the three types of second-order parts, become invalid and hence are removed. Second, for the approach to output valid parse trees during testing, we run maximum spanning tree (MST) \cite{mcdonald-etal-2005-non} based on the posterior edge probabilities predicted by the approach. 

Inspired by \citet{wang-etal-2019-second}, below we propose a \textbf{Local} second-order parsing approach.
While the \textbf{Single} approach uses Boolean random variables to represent existence of possible dependency edges, our \textbf{Local} approach defines a discrete random variable for each word specifying its dependency head, thus enforcing the head-selection constraint and leading to different formulation of the message passing inference steps.

\subsection{Scoring}
Following \citet{dozat2016deep}, we predict edge existence and edge labels separately. Suppose the input sentence is $\mathbf{w}=[w_0, w_1,w_2,\dots,w_n]$ where $w_0$ is a dummy root. We feed word representations outputted by the BiLSTM encoder into a biaffine function to assign score $s_{ij}^{\textrm{(edge)}}$ to edge $w_i \rightarrow w_j$. We use a Trilinear function to assign score $s_{ij,ik}^{\textrm{(sib)}}$ to the siblings part consisting of edges $w_i \rightarrow w_j$ and $w_i \rightarrow w_k$, and another Trilinear function to assign score $s_{ij,jk}^{\textrm{(gp)}}$ to the grandparent part consisting of edges $w_i \rightarrow w_j$ and $w_j \rightarrow w_k$. For edge labels, we use a biaffine function to predict label scores of each potential edge and use a softmax function to compute the label distribution $P(y_{ij}^{\text{(label)}}|\wvec)$, where $y_{ij}^{\text{(label)}}$ represents the possible label for edge $w_i \rightarrow w_j$.

\subsection{Message Passing}
The head-selection structured constraint requires that each word except the root has exactly one head.
We define variable $X_j\in\{0,1,2,\dots,n\}$ to indicate the head of word $w_j$. We then define a conditional random field (CRF) over $[X_1, \ldots, X_n]$. For each variable $X_j$, the unary potential is defined by:
\begin{align*}
    \phi_{u}(X_{j}=i)=&\exp(s_{ij}^{\textrm{(edge)}})
\end{align*}
Given two variables $X_j$ and $X_l$, the binary potential is defined by:
\begin{align*}
\phi_{p}(X_{j}=i,X_{l}=k)&=
\begin{cases}
\exp(s^{\text{(sib)}}_{ij,kl}) & k=i\\
\exp(s^{\text{(gp)}}_{ij,kl}) & k=j\\
1 & \text{Otherwise}
\end{cases}
\end{align*}
We use MFVI for approximate inference on this CRF. The algorithm updates the factorized posterior distribution $Q_{j}(X_j)$ of each word iteratively.
\begin{align*}
    \mathcal{M}^{(t-1)}_{j}(i)=&\sum_{k\neq i,j}Q^{(t-1)}_{k}(i)s^{(sib)}_{ij,ik}\\
    +&Q^{(t-1)}_{k}(j)s^{(gp)}_{ij,jk}+Q^{(t-1)}_{i}(k)s^{(gp)}_{ki,ij}\\
    Q_{j}^{(t)}(i) =&\frac{\mathrm{exp} \{s_{ij}^{\text{(edge)}}+\mathcal{M}^{(t-1)}_{j}(i)\}}{\sum\limits_{k=0}^n \mathrm{exp} \{s_{kj}^{\text{(edge)}}+\mathcal{M}^{(t-1)}_{j}(k)\}}     
\end{align*}
At $t=0$, $Q_{j}^{(t)}(X_j)$ is initialized by normalizing the unary potential.
The iterative update steps can be unfolded as recurrent neural network layers parameterized by part scores, thus forming an end-to-end neural network. 

Compared with the update formula in the \textbf{Single} approach, here the posterior distributions are defined over head-selections and are normalized over all possible heads. The computational complexity remains the same.

\subsection{Learning}
\label{sec:loss}
We define the cross entropy losses by:
\begin{align}
    \mathcal{L}^{\text{(edge)}}=&-\sum_{i}\log[Q_i(y_{i}^{*\textrm{(edge)}}|\mathbf{w})] \nonumber\\
    \mathcal{L}^{\textrm{(label)}}  =& -\sum_{i,j}\mathbbm{1}(y_{j}^{*\textrm{(edge)}}=i) \log(P (y_{ij}^{*\textrm{(label)}}|\mathbf{w}))\nonumber\\
    \mathcal{L} =& \lambda \mathcal{L}^{\text{(label)}} + (1-\lambda) \mathcal{L}^{\text{(edge)}}\nonumber
\end{align}
where $y_i^{*\textrm{(edge)}}$ is the head of word $w_i$ and $y_{ij}^{*\text{(label)}}$ is the label of edge $w_i\rightarrow w_j$ in the golden parse tree, $\lambda$ is a hyper-parameter and $\mathbbm{1}(x)$ is an indicator function that returns $1$ when $x$ is true and $0$ otherwise. 


\begin{table}[t!]
\small
\begin{center}
\begin{tabular}{lr}
\hline \hline
\textbf{Hidden Layer} & \textbf{Hidden Sizes}\\ \hline
Word/GloVe/Char & 100\\
POS & 50 \\
GloVe Linear & 125 \\
BERT Linear & 125 \\
BiLSTM & 3*600 \\
Char LSTM & 1*400 \\
Unary Arc (UD) & 500\\
\textbf{Local1O}/\textbf{Local2O} Unary Arc (Others) & 450\\
\textbf{Single1O}/\textbf{Single2O} Unary Arc (Others) & 550\\
Label & 150\\
Binary Arc & 150\\
\hline \textbf{Dropouts} & \textbf{Dropout Prob.}\\ \hline
Word/GloVe/POS & 20\%\\
Char LSTM (FF/recur) & 33\%\\
Char Linear & 33\%\\
BiLSTM (FF/recur) & 45\%/25\%\\
Unary Arc/Label & 25\%/33\%\\
Binary Arc & 25\%\\
\hline \textbf{Optimizer \& Loss} & \textbf{Value}\\ \hline
\textbf{Local1O}/\textbf{Local2O} Interpolation ($\lambda$)& 0.40\\
\textbf{Single1O}/\textbf{Single2O} Interpolation ($\lambda$)& 0.07\\
Adam $\beta_1$ & 0\\
Adam $\beta_2$ & 0.95\\
Decay Rate & 0.85 \\
Decay Step (without \textbf{dev} improvement) & 500 \\
\hline
\textbf{Weight Initialization} & \textbf{Mean/Stddev}\\
\hline
Unary weight & 0.0/1.0\\
Binary weight & 0.0/0.25\\
\hline \hline
\end{tabular}
\end{center}
\caption{Hyper-parameter for \textbf{Local1O}, \textbf{Single2O} and \textbf{Local2O} in our experiment. }
\label{tab:hyper}
\end{table}

\begin{table}[t!]
\centering
{
\small
\setlength\tabcolsep{4pt}
\begin{tabular}[t]{l|cc|cc}
\hlineB{4}
 & \multicolumn{2}{c|}{\textbf{PTB}} & \multicolumn{2}{c}{\textbf{CTB}}  \\
 
 & UAS & LAS & UAS & LAS  \\
\hline
\citet{dozat2016deep}                               & 95.74 & 94.08 & 89.30 & 88.23 \\
\citet{ma-etal-2018-stack}$^{\spadesuit}$                          & 95.87 & 94.19 & 90.59 & 89.29 \\
F\&G \shortcite{fernandez-gonzalez-gomez-rodriguez-2019-left}$^{\spadesuit}$ & 96.04 & 94.43 & - & -\\
GNN                                                                  & 95.87 & 94.15 & 90.78 & 89.50 \\
Single1O                                                            & 95.75 & 94.04 & 90.53 & 89.28 \\
Local1O                                                            & 95.83 & 94.23 & 90.59 & 89.28 \\
Single2O                                                           & 95.86 & 94.19 & 90.75 & 89.55 \\
Local2O                                                            & 95.98 & 94.34 & \textbf{90.81} & \textbf{89.57} \\
\hline
\citet{ji-etal-2019-graph}$^{\dagger}$                     & 95.97 & 94.31 & -     & -     \\
\citet{zhang2020efficient}$^{\dagger\ddagger}$                     & \textbf{96.14} & \textbf{94.49} & -     & -     \\
Local2O$^{\dagger\ddagger}$ & 96.12 & 94.47 & - & -\\
\hline\hline
\multicolumn{5}{c}{\textbf{+BERT}}\\
\hline
\citet{zhou-zhao-2019-head}$^{\clubsuit}$ & 97.20 & 95.72\\
\hline
\citet{clark-etal-2018-semi}$^{\diamond}$           & 96.60 & 95.00 & -     & -     \\
Single1O                                                            & 96.82 & 95.20 & 92.73 & 91.64 \\
Local1O                                                            & 96.86 & 95.32 & 92.47 & 91.30 \\
Single2O                                                           & 96.86 & 95.31 & \textbf{92.78} & \textbf{91.69} \\
Local2O                                                            & \textbf{96.91} & \textbf{95.34} & 92.55 & 91.38 \\
\hlineB{4}
\end{tabular}}
\caption{Comparison of our approaches and the previous state-of-the-art approaches on PTB and CTB. We report our results averaged over 5 runs. $^{\dagger}$: These approaches perform model selection based on the score on the development set. $^{\ddagger}$: These approaches do not use POS tags as input. $^{\diamond}$: \citet{clark-etal-2018-semi} uses semi-supervised multi-task learning with ELMo embeddings. $^{\spadesuit}$: These approaches use structured-skipgram embeddings instead of GloVe embeddings for PTB. $^{\clubsuit}$: For reference, \citet{zhou-zhao-2019-head} utilized both dependency and constituency information in their approach. Therefore, the results are not comparable to our results.} 
\label{tab:main:comparison}
\end{table}

\begin{table*}[t!]
\centering
\small
\setlength\tabcolsep{3.5pt}
\begin{tabular}{l||cccccccccccccc|c}
\hlineB{4}
	&	\textbf{PTB}	&	\textbf{CTB}	&	\textbf{bg}	&	\textbf{ca}	&	\textbf{cs}	&	\textbf{de}	&	\textbf{en}	&	\textbf{es}	&	\textbf{fr}	&	\textbf{it}	&	\textbf{nl}	&	\textbf{no}	&	\textbf{ro}	&	\textbf{ru}	&	\textbf{Avg.}	\\
\hline
GNN	&	94.15 	&	89.50\rlap{$^{\dagger}$}	&	90.33 	&	92.39 	&	90.95 	&	79.73 	&	88.43 	&	91.56 	&	87.23 	&	92.44 	&	88.57 	&	89.38 	&	85.26 	&	91.20 	&	89.37 	\\
Single1O	&	94.04 	&	89.28	&	90.05 	&	92.72\rlap{$^{\dagger}$}	&	92.07 	&	81.73 	&	89.55 	&	92.10 	&	88.27 	&	92.64 	&	89.57 	&	91.81 	&	85.39 	&	92.60 	&	90.13 	\\
Local1O	&	94.23 	&	89.28 	&	90.30 	&	92.56 	&	\textbf{92.15}	&	81.42 	&	89.43 	&	91.99 	&	88.26 	&	92.49 	&	89.76 	&	\textbf{91.91}	&	85.27 	&	\textbf{92.72}	&	90.13 	\\
Single2O	&	94.19 	&	89.55\rlap{$^{\dagger}$}	&	90.24 	&	92.82\rlap{$^{\dagger}$}	&	92.13 	&	\textbf{81.99\rlap{$^{\dagger}$}}	&	89.64\rlap{$^{\dagger}$}	&	\textbf{92.17\rlap{$^{\dagger}$}}	&	\textbf{88.69}	&	\textbf{92.83\rlap{$^{\dagger}$}}	&	89.97\rlap{$^{\dagger}$}	&	91.90 	&	85.53\rlap{$^{\dagger}$}	&	92.58 	&	90.30\rlap{$^{\dagger}$}	\\
Local2O	&	\textbf{94.34\rlap{$^{\dagger\ddagger}$}}	&	\textbf{89.57\rlap{$^{\dagger}$}}	&	\textbf{90.53\rlap{$^{\dagger}$}}	&	\textbf{92.83\rlap{$^{\dagger}$}}	&	92.12 	&	81.73 	&	\textbf{89.72\rlap{$^{\dagger}$}}	&	92.07 	&	88.53 	&	92.78 	&	\textbf{90.19\rlap{$^{\dagger}$}}	&	91.88 	&	\textbf{85.88\rlap{$^{\dagger\ddagger}$}}	&	92.67 	&	\textbf{90.35\rlap{$^{\dagger}$}}	\\
\hline
\hline
\multicolumn{16}{c}{\textbf{+BERT}}\\
\hline
Single1O	&	95.20 	&	91.64\rlap{$^{\dagger}$} 	&	90.87 	&	93.55\rlap{$^{\dagger}$}	&	92.01 	&	81.95\rlap{$^{\dagger}$}	&	90.44\rlap{$^{\dagger}$}	&	92.56\rlap{$^{\dagger}$}	&	\textbf{89.35} 	&	93.44\rlap{$^{\dagger}$}	&	90.89 	&	91.78 	&	86.13\rlap{$^{\dagger}$}	&	92.51 	&	90.88\rlap{$^{\dagger}$}	\\
Local1O	&	95.32 	&	91.30 	&	91.03 	&	93.17 	&	91.93 	&	81.66 	&	90.09 	&	92.32 	&	89.26 	&	93.05 	&	90.93 	&	91.62 	&	85.67 	&	92.51 	&	90.70 	\\
Single2O	&	95.31 	&	\textbf{91.69\rlap{$^{\dagger\ddagger}$}}	&	\textbf{91.30$^{\dagger}$}	&	\textbf{93.60\rlap{$^{\dagger\ddagger}$}}	&	\textbf{92.09\rlap{$^{\dagger}$}}	&	\textbf{82.00\rlap{$^{\dagger\ddagger}$}}	&	\textbf{90.75\rlap{$^{\dagger\ddagger}$}}	&	\textbf{92.62\rlap{$^{\dagger\ddagger}$}}	&	89.32	&	\textbf{93.66\rlap{$^{\dagger}$}}	&	\textbf{91.21}	&	\textbf{91.74}	&	\textbf{86.40\rlap{$^{\dagger}$}}	&	92.61 	&	\textbf{91.02\rlap{$^{\dagger\ddagger}$}}	\\
Local2O	&	\textbf{95.34}	&	91.38 	&	91.13 	&	93.34\rlap{$^{\dagger}$}	&	92.07\rlap{$^{\dagger}$}	&	81.67 	&	90.43\rlap{$^{\dagger}$}	&	92.45\rlap{$^{\dagger}$}	&	89.26 	&	93.50\rlap{$^{\dagger}$}	&	90.99 	&	91.66 	&	86.09\rlap{$^{\dagger}$}	&	\textbf{92.66}	&	90.86\rlap{$^{\dagger}$}\\
\hlineB{4}
\end{tabular}
\caption{LAS and standard deviations on test sets. We report results averaged over 5 runs. We use ISO 639-1 codes to represent languages from UD. $\dagger$ means that the model is statistically significantly better than the \textbf{Local1O} model by Wilcoxon rank-sum test with a significance level of $p<0.05$. We use $\ddagger$ to represent winner of the significant test between the \textbf{Single2O} and \textbf{Local2O} models.} 
\label{tab:lang_res}
\end{table*}

\section{Experiments}
\subsection{Setups}
\label{sec:setup}
Following previous work \cite{dozat2016deep,ma-etal-2018-stack}, we use PTB 3.0 \cite{marcus-etal-1993-building}, CTB 5.1 \cite{xue-etal-2002-building} and 12 languages in Universal Dependencies \cite{11234/1-2837} (UD) 2.2 to evaluate our parser. Punctuation is ignored in all the evaluations. We use the same treebanks and preprocessing as \citet{ma-etal-2018-stack} for PTB, CTB, and UD. For all the datasets, we remove sentences longer than 90 words in training sets for faster computation. 

We use \textbf{GNN}, \textbf{Local1O}, \textbf{Single1O}, \textbf{Local2O} and \textbf{Single2O} to represent the approaches of \citet{ji-etal-2019-graph}, \citet{dozat2016deep}, \citet{dozat-manning-2018-simpler}, and our two second-order approaches respectively. For all the approaches, we use the MST algorithm to guarantee tree-structured output in testing. We use the concatenation of word embeddings, character-level embeddings and part-of-speech (POS) tag embeddings to represent words and additionally concatenate BERT embeddings for experiments with BERT. For a fair comparison with previous work, we use GloVe \cite{pennington2014glove} and BERT-Large-Uncased model for PTB, and structured-skipgram \cite{ling-etal-2015-two} and BERT-Base-Chinese model for CTB. For UD, we use fastText embeddings \cite{bojanowski2017enriching} and BERT-Base-Multilingual-Cased model for different languages. We set the default iteration number for our approaches to 3 because we find no improvement on more or less iterations. 

For \textbf{GNN}\footnote{\url{https://github.com/AntNLP/gnn-dep-parsing}}, we rerun the code based on the official release of \citet{ji-etal-2019-graph}. For \textbf{Single1O}, \textbf{Local1O}\footnote{\url{https://github.com/tdozat/Parser-v3}}, \textbf{Single2O}\footnote{\url{https://github.com/wangxinyu0922/Second_Order_SDP}}, we implement these approaches based on the official release code of \citet{wang-etal-2019-second} and we implement \textbf{Local2O} based on this code. In speed comparison, we implement the second-order approaches based on an PyTorch implementation biaffine parser\footnote{\url{https://github.com/yzhangcs/parser}} implemented by \citet{zhang2020efficient} for a fair speed comparison with their approach\footnote{At the time we finished the paper, the official code for the second-order tree CRF parser have not release yet. We believe it is a fair comparison since we use the same settings and GPU as \citet{zhang2020efficient}.}. Since we find that the accuracy of our approaches based on PyTorch implementation on PTB does not change, we only report scores based on \citet{wang-etal-2019-second}.
\subsection{Hyper-parameters}

The hyper-parameters we used in our experiments is shown in Table \ref{tab:hyper}. We tune the the hidden size for calculating $s_{ij}^{\text{(edge)}}$ 
(Unary Arc in the table) separately for PTB and CTB. Following \citet{qi-etal-2018-universal}, we switch to AMSGrad \cite{j.2018on} after 5,000 iterations without improvement. We train models for 75,000 iterations with batch sizes of 6000 tokens and stopped the training early after 10,000 iterations without improvements on development sets. Different from previous approaches such as \citet{dozat2016deep} and \citet{ji-etal-2019-graph}, we use Adam \cite{kingma2014adam} with a learning rate of 0.01 and anneal the learning rate by 0.85 for every 500 iterations without improvement on the development set for optimization. For \textbf{GNN}, we train the models with the same setting as in \citet{ji-etal-2019-graph}. We do not use character embeddings and our optimization settings for \textbf{GNN} because we find they do not improve the accuracy. 

For the edge loss of \textbf{Single} approaches, \citet{zhang-etal-2019-empirical} proposed to sample a subset of the negative edges to balance positive and negative examples, but we find that using a relatively small interpolation $\lambda$ (shown in Table \ref{tab:hyper}) on label loss can improve the accuracy and the sampling does not help further improve the accuracy.

\subsection{Results}
Table \ref{tab:main:comparison} shows the Unlabeled Attachment Score (UAS) and Labeled Attachment Score (LAS) of all the approaches as well as the reported scores of previous state-of-the-art approaches on PTB and CTB. It can be seen that without BERT, our \textbf{Local2O} achieves state-of-the-art performance on CTB and has almost the same accuracy as the very recent work of \citet{zhang2020efficient} on PTB. With BERT embeddings, \textbf{Local2O} performs the best on PTB while \textbf{Single2O} has the best accuracy on CTB. 

Table \ref{tab:lang_res} shows the results of the five approaches on UD in addition to PTB and CTB. We make the following observations.
First, our second-order approaches outperform \textbf{GNN} and the first-order approaches both with and without BERT embeddings, showing that second-order decoders are still helpful in neural parsing even with strong contextual embeddings. Second, without BERT, \textbf{Local} slightly outperforms \textbf{Single}, although the difference between the two is quite small\footnote{Note that \citet{zhang-etal-2019-empirical} reports higher difference in accuracy between first-order \textbf{Local} and \textbf{Single} approaches. The discrepancy is most likely caused by our better designed loss function and tuned hyper-parameters.}; when BERT is used, however, \textbf{Single} clearly outperforms \textbf{Local}, which is quite interesting and warrants further investigation in the future. 
Third, the relative strength of \textbf{Local} and \textbf{Single} approaches varies over treebanks, suggesting varying importance of the head-selection constraint. 


\begin{table}[t!]
\centering
\small
\setlength\tabcolsep{3.5pt}
\begin{tabular}{l|ccc}
\hlineB{4}
System & Train & Test & Time Complexity\\
\hline
GNN & 392 & 464 & $O(n^2d)$\\
\citet{zhang2020efficient} & 200 & 400 & $O(n^3)$\\
Single1O & 616 & 1123 & $O(n^2)$ \\
Local1O & \textbf{625} & \textbf{1150} & $O(n^2)$ \\
Single2O & 481 & 966 & $O(n^3)$ \\
Local2O & 486 & 1006 & $O(n^3)$ \\
\hlineB{4}
\end{tabular}
\caption{Comparison of training and testing speed (sentences per second) and the time complexity of the decoders of different approaches on PTB. }
\label{tab:speed}
\end{table}

\subsection{Speed Comparison}
We evaluate the speed of different approaches on a single GeForce GTX 1080 Ti GPU following the setting of \citet{zhang2020efficient}. As shown in Table \ref{tab:speed}, our \textbf{Local} approach and \textbf{Single} approach have almost the same speed. Our second-order approaches only slow down the training and testing speed in comparison with the first-order approaches by 23\% and 12\% respectively. 
They are also significantly faster than previous state-of-the-art approaches. Our \textbf{Local} approach is 1.2 and 2.3 times faster than \textbf{GNN} in training and testing respectively and is 2.4 and 2.9 times faster than the second-order tree CRF approach of \citet{zhang2020efficient}. 

In terms of time complexity, our second-order decoders have a time complexity of $O(n^3)$\footnote{The MST algorithm has a time complexity of $O(n^2)$ and we follow \citet{dozat-etal-2017-stanfords} only using the MST algorithm when the argmax predictions of structured output are not trees.}; while the time complexity of \textbf{GNN} is $O(n^2d)$, the hidden size $d$ (500 by default) is typically much larger than sentence length $n$; and the decoder of \citet{zhang2020efficient} has a time complexity of $O(n^3)$ as well, but it requires sequential computation over the input sentence while our decoders can be parallelized over words of the input sentence. 

\section{Conclusion}
We propose second-order graph-based dependency parsing based on message passing and end-to-end neural networks. We modify a previous approach that predicts dependency edges independently and also design a new approach that incorporates the head-selection structured constraint. Our experiments show that our second-order approaches have better overall performance than the first-order baselines; they achieve competitive accuracy with very recent start-of-the-art second-order graph-based parsers and are significantly faster. Our empirical comparisons also show that second-order decoders still outperform first-order decoders even with BERT embeddings, and that the usefulness of the head-selection constraint is limited, especially when using BERT embeddings. Our code is publicly available at \url{https://github.com/wangxinyu0922/Second_Order_Parsing}.

\section*{Acknowledgements}
This work was supported by the National Natural Science Foundation of China (61976139).

\bibliography{anthology,aacl-ijcnlp2020}
\bibliographystyle{acl_natbib}

\end{document}